# Stress Detection on Code-Mixed Texts in Dravidian Languages using Machine Learning

**L. Ramos, M. Shahiki-Tash, Z. Ahani, A. Eponon, O. Kolesnikova, H. Calvo**
Instituto Politécnico Nacional, Centro de Investigación en Computación
Mexico City, Mexico

Corresponding author: Moein Shahiki Tash
`mshahikit2022@cic.ipn.mx`

**Abstract**

Stress is a common feeling in daily life, but it can affect mental well-being in some situations, the development of robust detection models is imperative. This study introduces a methodical approach to the stress identification in code-mixed texts for Dravidian languages. The challenge encompassed two datasets, targeting Tamil and Telugu languages respectively. This proposal underscores the importance of using uncleaned text as a benchmark to refine future classification methodologies, incorporating diverse preprocessing techniques. Random Forest algorithm was used, featuring three textual representations: TF-IDF, Uni-grams of words, and a composite of (1+2+3)-Grams of characters. The approach achieved a good performance for both linguistic categories, achieving a Macro F1-score of 0.734 in Tamil and 0.727 in Telugu, overpassing results achieved with different complex techniques such as FastText and Transformer models. The results underscore the value of uncleaned data for mental state detection and the challenges classifying code-mixed texts for stress, indicating the potential for improved performance through cleaning data, other preprocessing techniques or more complex models.

## 1 Introduction

According to the World Health Organization (WHO)[1], stress is characterized as a condition of anxiety or mental strain caused by challenging circumstances. Every individual experiences a certain stress level, as it is an inherent reaction to threats and various stimuli. Thus, not all stress states are harmful; chronicity, quality, magnitude, subjective appraisal, and context of stressors are important moderators of the stress response, but acute and chronic stress experiences can affect optimal neuroendocrine reactivity, leading to increased vulnerability of the organism to stressors(Agorastos and Chrousos, 2022). Social media is considered as a platform where users express themselves. The rise of social media as one of humanity's most important public communication platforms presents a potential prospect for early identification and management of mental illness(Andrew, 2024). Developing resilient methods for promptly identifying human stress is crucial, and these technologies offer the potential for ongoing stress monitoring(Li and Liu, 2020). The prevalence of multilingualism on the internet, and code-mixed text data, has become a popular research topic in natural language processing (NLP). It is a difficult task to handle bilingual and multilingual communication data.(Yigezu et al., 2022), one of these tasks is hope speech detection(Ahani et al., 2024c; Tash et al., 2024b; Arif et al., 2024) or hate speech detection(Zamir et al., 2024; Ahani et al., 2024a; Tash et al., 2024a). Texts in mixed language pose a significant challenge. Numerous users seek a straightforward method to form sentences or use familiar expressions. They attempt to compose a text that combines two or three different languages, resulting in the generation of Code-Mix data (Tash et al., 2022). Some applications have been developed using NLP and ML in Dravidian languages, such as sentiment classification(Rashmi et al., 2021), abuse detection (Bansal et al., 2022), hate and offensive content identification (Rajalakshmi et al., 2023) or hate and offensive language detection (Roy et al., 2022). There have not been any recent attempts to identify stress in Dravidian languages like Tamil and Telugu. This approach seeks to bridge the gap by addressing the task with foundational methods that utilize text representation models, bypassing preprocessing steps such as the deletion of stop words, special characters, emojis, spelling symbols, etc. The results represent a benchmark for assessing future applications that employ diverse techniques in text preprocessing, feature extraction and ML models.

---
[1]https://www.who.int/news-room/questions-and-answers/item/stress

The remaining sections of this paper are structured as follows: Section 2 contains the related works about stress detection using ML techniques. Section 3 presents the task description. Section 4 presents a data description. Section 5 makes a summary of the methodology followed. Section 6 presents the results and discussion, and the study's conclusions are presented in Section 7. Finally, Section 8 presents the limitations of this approach.

| Data set | Label | Train | Validation | Test | Total |
|---|---|---|---|---|---|
| Tamil | Non stressed | 3720 | 939 | 650 | 5309 |
| | stressed | 1784 | 439 | 370 | 2593 |
| Telugu | Non stressed | 3314 | 799 | 650 | 4763 |
| | stressed | 1783 | 440 | 400 | 2623 |

Table 1: Data set distribution

| Data set | Label | Text |
|---|---|---|
| Tamil | Non stressed | Bro video clip swap agi iruku atha gavanichingala |
| | stressed | bhaLLi Suttu vīzhthappattadhu. |
| Telugu | Non stressed | super comment pettav bro , chala navvostundi |
| | stressed | Nēnu 10 rōjulaṅgā snānam chēyalēdu! |

Table 2: Sample instances of data set

## 2 Related Work

Stress identification has been explored using some ML methods. (Nijhawan et al., 2022) employed sentiment analysis with five different labels (Joy, Sadness, Neutral, Anger, and Fear), Latent Dirichlet Allocation (LDA), and ML to detect mental stress in social media texts, obtained their best results using Random Forest (RF) with a precision of 97.78%. Yang et al. collected texts from Twitter (today X) and applied them to filter using predefined patterns to obtain data related to stresses, performed manual tagging and used several classifiers. Lowercasing and anonymizing URLs and usernames was the preprocessing process for each text. The 20,000 most frequent N-Grams were used as text representation, each word or character sequence was replaced with a dense numerical vector. Their best result was obtained with BERT reporting an F1-score of 0.86 for the negative class and 0.79 for the positive class and an accuracy of 83.6In shared task, various teams employed diverse methodologies to tackle the detection of hope speech across both tasks. Similar to the previous one, teams utilized a range of machine learning approaches, including traditional classifiers like LR, SVM and advanced deep learning techniques(Ahani et al., 2024b) such as Transformers and LLM-based models. The StressIdent‿LT-EDI@EACL2024[2] shared task had the aim to detect whether a person is affected by stress from their social media postings wherein people share their feelings and emotions(Tash et al., 2024c). Given social media postings in Tamil and Telugu code-mixed languages, the submitted system should classify into two labels, "stressed" or "not stressed". Teams utilized diverse machine learning approaches, including traditional classifiers like Logistic Regression, Support Vector Machine and deep learning techniques such as CNN, Transformers and LLM-based models. In this workshop, (Eponon et al., 2024) proposed an approach for stress detection in Tamil and Telugu languages achieved a macro F1 score of 0.77 for Tamil and 0.72 for Telugu with FastText and Naïve Bayes. (Raihan et al., 2024) tested several methods to classify stress in Tamil and Telugu languages, from traditional ML models to Transformers, but the best results were obtained using BERT-based models, these models achieved a macro F1 score of f 0.71 for Tamil and 0.72 for Telugu. (Andrew, 2024) uses GPT2 to detect stress in Tamil and Telugu languages. Although they used a transformer model with billions of parameters, it was not possible to achieve good performance. They only achieve a macro F1 score of 0.273 for Tamil and 0.251 for Telugu.

## 3 Data Description

The data that was used for this task was two datasets for educational purposes, and it was generously provided by the organizers of LT-EDI workshop. The first was the Tamil dataset and the second was the Telugu dataset, each one had two labels, "Non stressed" and "stressed" and was divided into the train, validation, and test datasets. In Table 1 it is observable the data set distribution for both languages. In addition, Table 2 shows some samples from the Tamil and Telugu dataset.

---
[2]https://codalab.lisn.upsaclay.fr/competitions/16092

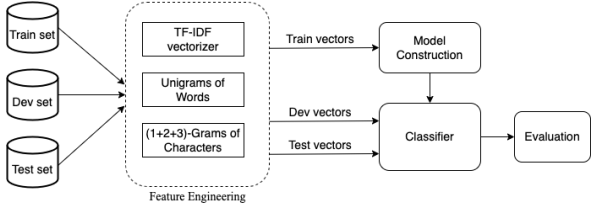

Figure 1: Overview of the proposed methodology

## 4 Methodology

This section outlines the methodology for each task. The primary objective of this proposal is to establish a baseline model that utilizes text representation models with no cleaning steps before the use of text representation models and employs the RF algorithm for classification. RF was selected due to the distribution of the random vectors does not depend on the training set, does not concentrate weight on any subset of the instances and the noise effect is smaller (Breiman, 2001). The overview of the proposed methodology is exposed in Figure 1.

### 4.1 Feature Engineering

In this approach, two vectorization methods have been used for each task, TF-IDF and N-Grams of words and characters. TF-IDF was selected due to TF-IDF considers every word's weight by using two approaches, the frequency of a term and in how many file a term can be found (Hakim et al., 2014), The term N-Gram (of characters or words) refers to a series of sequential tokens in a sentence, paragraph, and document, in addition the Uni-grams have shown good results in code-mixed text classification (Ameer et al., 2022), and Character N-Grams are handcrafted features which widely serve as discriminative features in text categorization, discriminating language variety, and many other applications (Kruczek et al., 2020). For creating these representations, the scikit-learn library has been used with its default parameters in "TfidfVectorizer" method and specifically set "ngram_range=(1,1)" for Uni-grams of words and "ngram_range=(1,3)" and "analyzer='char'" for (1+2+3)-Grams of characters and all others parameters in N-Grams methods remained in their default settings.

### 4.2 Evaluation

Standard metrics such as Macro F1-score, Macro Recall, Macro Precision, Weighted F1-score, Weighted Recall, Weighted Precision, and Accuracy were used to evaluate the performance.

## 5 Results and Discussion

The classification was performed using RF. The training was performed with the training set and after an evaluation was performed using the validation set, the evaluation over the validation data set is shown in Table 3 and finally, predictions were made with the test set. The text was represented using three different vectorizations: TF-IDF, Uni-grams of words, and (1+2+3)-Grams of characters were selected, but the highest performance was obtained with Uni-grams of words in Tamil and for Telugu was TF-IDF. Table 4 shows the difference with the highest F1-score reported. Based on the results obtained, since the best representation of the text is different in each task, which suggests that for Tamil it is convenient a representation that consider individual words, while for Telugu the weighting of the term frequency-inverse frequency of documents was more effective, this suggests that it is due to the language characteristics. Furthermore, the results of the Macro F1-score, Macro Recall and Macro Precision metrics suggest that the model is not biased toward any specific category and reflecting the model's ability to generalize well.

## 6 Conclusion

In conclusion, this approach has proven to be able to identify stresses in texts using the RF algorithm without applying any cleaning techniques to the text, this is valuable as it allows having a baseline for future approaches in stress identification in Tamil and Telugu, especially when considering more complex methods such as Deep Learning methods and also the importance of the uncleaned data in code-mixed texts and detecting mental states. The results of this approach reveal the difficulties in classifying code-mixed texts in Dravidian languages such as Tamil and Telugu due to the combination of languages. In addition, this paper contributes to the growth of research in stress identification in code-mixed texts with valuable information for future research.

## 7 Limitations

This stress identification approach in code-mixed texts in Dravidian languages has demonstrated promising performance in these experiments, but it is essential to acknowledge certain limitations that warrant consideration. A notable constraint is the small amount of data that is provided, in addition to the imbalance that exists between classes. Therefore, the model may encounter challenges or facilities using other classification models and some preprocessing techniques, because this approach has no information about the impact of applying or not applying certain preprocessing techniques on classification performance. Addressing these limitations is important for further enhancing stress identification in code-mixed texts in Tamil and Telugu languages.

| Data set | Feature | Macro F1-score | Macro Recall | Macro Precision | Weighted F1-score | Weighted Recall | Weighted Precision | Accuracy |
|---|---|---|---|---|---|---|---|---|
| **Tamil** | TF-IDF | 0.979 | 0.974 | 0.984 | 0.982 | 0.982 | 0.982 | 0.982 |
| | Uni-grams of Words | 0.976 | 0.971 | 0.982 | 0.980 | 0.980 | 0.980 | 0.980 |
| | (1+2+3)-Grams of Characters | 0.997 | 0.997 | 0.998 | 0.998 | 0.998 | 0.998 | 0.998 |
| **Telugu** | TF-IDF | 0.992 | 0.992 | 0.992 | 0.993 | 0.993 | 0.993 | 0.993 |
| | Uni-grams of Words | 0.991 | 0.991 | 0.992 | 0.992 | 0.992 | 0.992 | 0.992 |
| | (1+2+3)-Grams of Characters | 0.994 | 0.994 | 0.994 | 0.994 | 0.994 | 0.994 | 0.994 |

Table 3: Random Forest performance with development data

| Data set | Author | Macro F1-score | Macro Recall | Macro Precision | Weighted F1-score | Weighted Recall | Weighted Precision | Accuracy |
|---|---|---|---|---|---|---|---|---|
| Tamil | (Eponon et al., 2024) | 0.725 | 0.775 | - | - | - | 0.822 | 0.724 |
| Telugu | | 0.727 | 0.756 | - | - | - | 0.779 | 0.729 |
| Tamil | (Raihan et al., 2024) | 0.71 | - | - | - | - | - | 0.71 |
| Telugu | | 0.72 | - | - | - | - | - | 0.72 |
| Tamil | (Andrew, 2024) | 0.275 | 0.498 | 0.459 | 0.202 | 0.364 | 0.485 | - |
| Telugu | | 0.251 | 0.247 | 0.255 | 0.287 | 0.281 | 0.293 | - |
| Tamil | This proposal | 0.734 | 0.780 | 0.767 | 0.737 | 0.734 | 0.818 | 0.734 |
| Telugu | | 0.727 | 0.763 | 0.755 | 0.730 | 0.728 | 0.794 | 0.728 |

Table 4: Performance of the model proposed in comparison with other proposals

## Acknowledgments

The work was done with partial support from the Mexican Government through the grant A1-S-47854 of CONACYT, Mexico, grants 20241816, 20241819, and 20240951 of the Secretaría de Investigación y Posgrado of the Instituto Politécnico Nacional, Mexico. The authors thank the CONACYT for the computing resources brought to them through the Plataforma de Aprendizaje Profundo para Tecnologías del Lenguaje of the Laboratorio de Supercómputo of the INAOE, Mexico and acknowledge the support of Microsoft through the Microsoft Latin America PhD Award.

## References

Agorastos Agorastos and George P Chrousos. 2022. The neuroendocrinology of stress: the stress-related continuum of chronic disease development. *Molecular Psychiatry*, 27(1):502–513.

Z Ahani, M Tash, M Zamir, and I Gelbukh. 2024a. Zavira@ dravidianlangtech 2024: Telugu hate speech detection using lstm. In *Proceedings of the Fourth Workshop on Speech, Vision, and Language Technologies for Dravidian Languages*, pages 107–112.

Zahra Ahani, Moein Shahiki Tash, Yoel Ledo Mezquita, and Jason Angel. 2024b. Utilizing deep learning models for the identification of enhancers and super-enhancers based on genomic and epigenomic features. *Journal of Intelligent & Fuzzy Systems*, (Preprint):1–11.

Zahra Ahani, Moein Shahiki Tash, Majid Tash, Alexander Gelbukh, and Irna Gelbukh. 2024c. Multiclass hope speech detection through transformer methods. In *Proceedings of the Iberian Languages Evaluation Forum (IberLEF 2024), co-located with the 40th Conference of the Spanish Society for Natural Language Processing (SEPLN 2024), CEUR-WS. org*.

Iqra Ameer, Grigori Sidorov, Helena Gomez-Adorno, and Rao Muhammad Adeel Nawab. 2022. Multi-label emotion classification on code-mixed text: Data and methods. *IEEE Access*, 10:8779–8789.

Judith Jeyafreeda Andrew. 2024. Judithjeyafreeda stressident lt-edi@ eacl2024: Gpt for stress identification. In *Proceedings of the Fourth Workshop on Language Technology for Equality, Diversity, Inclusion*, pages 173–176.

Muhammad Arif, Moein Shahiki Tash, Ainaz Jamshidi, Iqra Ameer, Fida Ullah, Jugal Kalita, Alexander Gelbukh, and Fazlourrahman Balouchzahi. 2024. Exploring multidimensional aspects of hope speech computationally: A psycholinguistic and emotional perspective. Preprint.

Vibhuti Bansal, Mrinal Tyagi, Rajesh Sharma, Vedika Gupta, and Qin Xin. 2022. A transformer based approach for abuse detection in code mixed indic languages. *ACM transactions on Asian and low-resource language information processing*.

Leo Breiman. 2001. Random forests. *Machine learning*, 45:5–32.

Anvi Alex Eponon, Ildar Batyrshin, and Grigori Sidorov. 2024. Pinealai stressident_lt-edi@eacl2024: Minimal configurations for stress identification in tamil and telugu. In *Proceedings of the Fourth Workshop on Language Technology for Equality, Diversity, Inclusion*, pages 152–156.

Ari Aulia Hakim, Alva Erwin, Kho I Eng, Maulahikmah Galinium, and Wahyu Muliady. 2014. Automated document classification for news article in bahasa indonesia based on term frequency inverse document frequency (tf-idf) approach. In *2014 6th international conference on information technology and electrical engineering (ICITEE)*, pages 1–4. IEEE.

Jakub Kruczek, Paulina Kruczek, and Marcin Kuta. 2020. Are n-gram categories helpful in text classification? In *Computational Science–ICCS 2020: 20th International Conference, Amsterdam, The Netherlands, June 3–5, 2020, Proceedings, Part II 20*, pages 524–537. Springer.

Russell Li and Zhandong Liu. 2020. Stress detection using deep neural networks. *BMC Medical Informatics and Decision Making*, 20:1–10.

Tanya Nijhawan, Girija Attigeri, and T Ananthakrishna. 2022. Stress detection using natural language processing and machine learning over social interactions. *Journal of Big Data*, 9(1):33.

Abu Raihan, Tanzim Rahman, Md Rahman, Jawad Hossain, Shawly Ahsan, Avishek Das, and Mohammed Moshiul Hoque. 2024. Cuet_duo@stressident_lt-edi@eacl2024: Stress identification using tamil-telugu bert. In *Proceedings of the Fourth Workshop on Language Technology for Equality, Diversity, Inclusion*, pages 265–270.

Ratnavel Rajalakshmi, Srivarshan Selvaraj, Pavitra Vasudevan, et al. 2023. Hottest: Hate and offensive content identification in tamil using transformers and enhanced stemming. *Computer Speech & Language*, 78:101464.

KB Rashmi, HS Guruprasad, and BR Shambhavi. 2021. Sentiment classification on bilingual code-mixed texts for dravidian languages using machine learning methods. In *FIRE (Working Notes)*, pages 899–907.

Pradeep Kumar Roy, Snehaan Bhawal, and Chinnaudayar Navaneethakrishnan Subalalitha. 2022. Hate speech and offensive language detection in dravidian languages using deep ensemble framework. *Computer Speech & Language*, 75:101386.

M Shahiki Tash, Z Ahani, Al Tonja, M Gemeda, N Hussain, and O Kolesnikova. 2022. Word level language identification in code-mixed kannada-english texts using traditional machine learning algorithms. In *Proceedings of the 19th International Conference on Natural Language Processing (ICON): Shared Task on Word Level Language Identification in Code-mixed Kannada-English Texts*, pages 25–28.

M Tash, Z Ahani, M Zamir, O Kolesnikova, and G Sidorov. 2024a. Lidoma@ lt-edi 2024: Tamil hate speech detection in migration discourse. In *Proceedings of the Fourth Workshop on Language Technology for Equality, Diversity, Inclusion*, pages 184–189.

Moein Shahiki Tash, Zahra Ahani, Mohim Tash, Olga Kolesnikova, and Grigori Sidorov. 2024b. Exploring sentiment dynamics and predictive behaviors in cryptocurrency discussions by few-shot learning with large language models. *arXiv preprint arXiv:2409.02836*.

Moein Shahiki Tash, Olga Kolesnikova, Zahra Ahani, and Grigori Sidorov. 2024c. Psycholinguistic and emotion analysis of cryptocurrency discourse on x platform. *Scientific Reports*, 14(1):8585.

Mesay Gemeda Yigezu, Atnafu Lambebo Tonja, Olga Kolesnikova, Moein Shahiki Tash, Grigori Sidorov, and Alexander Gelbukh. 2022. Word level language identification in code-mixed kannada-english texts using deep learning approach. In *Proceedings of the 19th International Conference on Natural Language Processing (ICON): Shared Task on Word Level Language Identification in Code-mixed Kannada-English Texts*, pages 29–33.

Muhammad Zamir, Moein Tash, Zahra Ahani, Alexander Gelbukh, and Grigori Sidorov. 2024. Lidoma@ dravidianlangtech 2024: Identifying hate speech in telugu code-mixed: A bert multilingual. In *Proceedings of the Fourth Workshop on Speech, Vision, and Language Technologies for Dravidian Languages*, pages 101–106.